\documentclass[11pt]{article}

\usepackage[preprint]{acl}

\usepackage{times}
\usepackage{latexsym}

\usepackage[T1]{fontenc}
\usepackage[utf8]{inputenc}

\usepackage{microtype}

\usepackage{inconsolata}

\usepackage{graphicx}
\usepackage{amsmath}
\usepackage{amssymb}
\usepackage{algorithm}
\usepackage{algpseudocode}
\newtheorem{lemma}{Lemma}
\usepackage{multirow}
\usepackage{tabularx}
\usepackage{booktabs}
\usepackage{pifont}
\usepackage[table]{xcolor}
\usepackage[symbol]{footmisc}
\usepackage{newfloat}
\usepackage{hyperref}

\newenvironment{links}{%
  \newcommand{\link}[2]{\par\textbf{##1} --- \url{##2}}%
  \setlength{\hangindent}{10pt}%
  \setlength{\parskip}{2pt}%
  \begin{flushleft}%
}{%
  \end{flushleft}%
  \vskip 1ex%
}
\title{FoRA: Fisher-orthogonal Rank Adaptation\\for Parameter-Efficient Fine-Tuning}

\author{
 \textbf{Juneyoung Park\textsuperscript{1}},
 \textbf{Seongbae Lee\textsuperscript{1}},
 \textbf{Han-Sang Lee\textsuperscript{2}},
 \textbf{Kyuho Lee\textsuperscript{2}},
 \textbf{Minjae Kim\textsuperscript{2}},
\\
 \textbf{Seungheon Hyeon\textsuperscript{2}\thanks{
 \raggedright
 Corresponding authors:
 Seungheon Hyeon,
 KIDUK KWON,
 Seongwan Kim,
 Jaeho Lee.
\\[2mm]
 \textsuperscript{1}OptAI Inc.:
 \{jyoung.park, sbae.lee, swan.kim, jaeho.lee\}@opt-ai.kr
\\
 \textsuperscript{2}LG Uplus:
 \{hansanglee, kyuholee, minjaekim, sheon, kwonkiduk\}@lguplus.co.kr
 }},
 \textbf{KIDUK KWON\textsuperscript{2}\footnotemark[1]},
 \textbf{Seongwan Kim\textsuperscript{1}\footnotemark[1]},
 \textbf{Jaeho Lee\textsuperscript{1}\footnotemark[1]}
\\
\\
 \textsuperscript{1}OptAI Inc,
 \textsuperscript{2}LG Uplus
}

\begin{document}
\maketitle
\begin{abstract}
Parameter-efficient fine-tuning(PEFT) has largely focused on LoRA and its accuracy-oriented variants, leaving the original goal of reducing trainable parameters has received comparatively little attention. We introduce FoRA, which revisits this goal by reducing the number of adapted layers rather than adapter rank. FoRA selects task-informative layers via a single-pass diagonal Fisher score (under 1\% of training cost) and trains the LoRA down-projection at selected layers on the Stiefel manifold, preserving column orthonormality and effective rank. FoRA consistently outperforms LoRA and DoRA at half their parameter budget, and falls within 0.7--0.8 accuracy points of AdaLoRA at one-quarter its parameter count, across five LLaMA-family backbones. Cross-architecture experiments on twelve backbones from the LLaMA, Qwen3, and Gemma families confirm consistent gains from 270M to 32B parameters. The two components combine super-additively: Fisher selection alone matches rank reduction at the same budget, while the Stiefel constraint provides the decisive additional gain.
\end{abstract}
\begin{links}
    \link{Code}{https://github.com/crinex/FoRA}
    % \link{Extended version}{https://arxiv.org/abs/}
\end{links}
\section{Introduction}

LoRA \citep{hu2022lora} has become the de facto standard for parameter-efficient fine-tuning (PEFT) by freezing the pretrained weights and learning only a small low-rank update $\Delta W = BA$. Subsequent work, including DoRA \citep{liu2024dora}, rsLoRA \citep{kalajdzievski2023rslora}, and PiSSA \citep{meng2024pissa}, has focused on improving accuracy or training dynamics under a fixed parameter budget. Recent studies report that the effective rank of trained LoRA adapters often falls well below the nominal rank \citep{biderman2024lora,hayou2024lora}, suggesting that gains along this accuracy-oriented axis are narrowing.

The original goal of LoRA, parameter efficiency itself, has received comparatively little attention. Reducing parameters in the LoRA family typically means lowering the rank $r$. This directly shrinks the subspace each adapter can represent, losing expressiveness at the very layers where adaptation is most needed. Layer-level methods such as LISA \citep{pan2024lisa} and LoRA-drop \citep{zhou2024loradrop} attempt selective adaptation, but rely on stochastic layer sampling or post-hoc pruning, both of which add overhead or require a full training pass before selection is fixed. AdaLoRA \citep{zhang2023adalora} redistributes per-layer rank dynamically but recomputes SVD-based importance at every step and introduces multiple schedule hyperparameters.

We take a different approach: rather than cutting the rank, we reduce the number of adapted layers while preserving per-adapter rank intact. Adapting only a small set of information-rich layers yields comparable or better performance at half the parameter count. We identify these layers using an empirical diagonal Fisher score computed in a single forward-backward pass before fine-tuning begins (below 1\% of training cost) and hold the selection fixed throughout training.

Reducing the layer count also concentrates task pressure on the remaining adapters. To ensure each adapter fully uses its capacity, we constrain the LoRA down-projection $B$ to the Stiefel manifold of column-orthonormal matrices via Cayley parametrization \citep{wen2013feasible}. This enforces that all $r$ orthogonal directions of each adapter are utilized, preventing the spectral collapse reported for unconstrained LoRA \citep{biderman2024lora,hayou2024lora}.

We combine these two ideas into \textbf{FoRA} (\textbf{F}isher-\textbf{o}rthogonal \textbf{R}ank \textbf{A}daptation): Fisher decides \emph{where} to adapt, and the Stiefel constraint shapes \emph{how} that capacity is used. The two components are orthogonal in design and combine super-additively, as our ablation confirms.

Our contributions are: (i) a static Fisher-based layer selection criterion that halves the adapter layer count at below 1\% calibration cost, enabling FoRA to outperform LoRA and DoRA at half their parameter budget and match AdaLoRA within 0.7--0.8 accuracy points at one-quarter its budget; (ii) a Stiefel-constrained adapter that restores effective rank from 0.71 to 0.88 of the nominal rank and combines super-additively with layer selection; and (iii) consistent validation across twelve backbones from the LLaMA, Qwen3, and Gemma families at scales from 270M to 32B parameters.

\begin{figure}[t]
\centering
\includegraphics[width=\columnwidth]{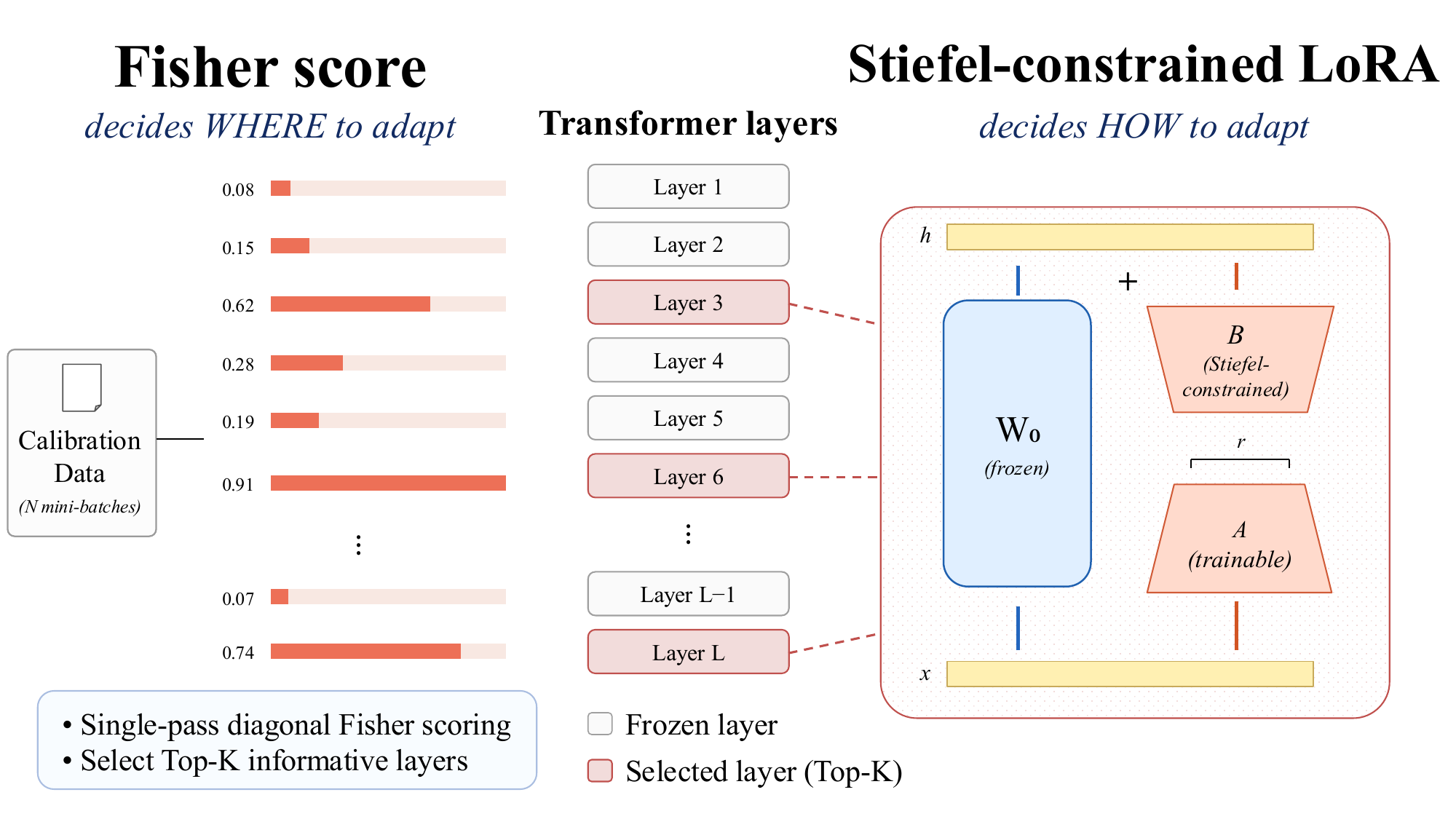}
\caption{Overview of FoRA. FoRA uses calibration data to score Transformer layers with diagonal Fisher information, selects the Top-K informative layers for adaptation, and applies Stiefel-constrained LoRA only to those layers while keeping the rest frozen.}
\label{fig:overview}
\end{figure}

\section{Related Work}

\noindent\textbf{LoRA and its variants.}
LoRA \citep{hu2022lora} freezes the pretrained weight $W_0$ and learns a low-rank update $\Delta W = BA$. Follow-up methods improve accuracy or training stability at a fixed parameter budget: DoRA \citep{liu2024dora} decomposes updates into magnitude and direction; rsLoRA \citep{kalajdzievski2023rslora} corrects rank scaling; PiSSA \citep{meng2024pissa} initializes from the dominant singular vectors; LoRA+ \citep{hayou2024lora} uses asymmetric learning rates for $A$ and $B$. These methods can be combined with base-model quantization, as in QLoRA \citep{dettmers2023qlora}. Despite these improvements, the effective rank of trained adapters is often substantially below the nominal rank $r$ \citep{biderman2024lora,hayou2024lora}, limiting further gains along this axis.

\vspace{0.5em}
\noindent\textbf{Selective layer-level fine-tuning.}
AdaLoRA \citep{zhang2023adalora} dynamically redistributes per-layer rank using SVD-based importance scores recomputed at every step. LISA \citep{pan2024lisa} samples a random layer subset at each iteration, and LoRA-drop \citep{zhou2024loradrop} prunes adapters post-hoc by output magnitude. All three determine the layer set during or after training, incurring calibration overhead or requiring a full pass before selection is fixed. FoRA instead computes a Fisher score in a single forward-backward pass before fine-tuning begins and holds the selection static throughout training.

\vspace{0.5em}
\noindent\textbf{Orthogonal and manifold constraints.}
OFT \citep{qiu2023oft} and BOFT \citep{liu2024boft} constrain weight updates to orthogonal transformations that preserve hyperspherical energy between neuron activations. VeRA \citep{kopiczko2024vera} takes an extreme compression approach by sharing a single pair of fixed random matrices across all layers and learning only per-layer scaling vectors, locking each adapter to a fixed random subspace while minimizing trainable parameters. In the LoRA family specifically, \citet{park2025riemannian} optimize the down-projection $B$ on the Stiefel manifold of column-orthonormal matrices via Cayley parametrization \citep{wen2013feasible}, showing that this constraint prevents spectral collapse and restores the effective rank of the adapter output. 

Consequently, FoRA integrates static Fisher-based layer selection with Stiefel-constrained down-projections, efficiently optimizing both adapter placement and rank utilization simultaneously.

\section{Method}
\label{sec:method}
\noindent\textbf{Preliminaries.}
We use the standard LoRA \citep{hu2022lora} parametrization. For a pretrained linear weight $W_0 \in \mathbb{R}^{d_\text{out} \times d_\text{in}}$ in a transformer layer, LoRA freezes $W_0$ and adds a low-rank update
\begin{equation}
W = W_0 + \Delta W, \qquad \Delta W = B A,
\label{eq:lora}
\end{equation}
where $A \in \mathbb{R}^{r \times d_\text{in}}$, $B \in \mathbb{R}^{d_\text{out} \times r}$, and $r \ll \min(d_\text{out}, d_\text{in})$. We denote the set of $L$ transformer layers by $\{\ell_1, \ldots, \ell_L\}$ and the union of trainable adapter parameters at layer $\ell$ by $\theta_\ell = \{A_\ell, B_\ell\}$ across all target modules. Following \citet{hu2023llmadapters}, the target modules at each adapted layer are the three self-attention projections $\{q, k, v\}$ and the two MLP projections $\{\text{up}, \text{down}\}$, for a total of five projections per layer. Standard LoRA applies adapters to every layer, giving a trainable parameter count of $L \cdot M \cdot r (d_\text{in} + d_\text{out})$ with $M$ target modules per layer. Our goal is to reduce the number of layers that carry adapters while preserving the per-adapter capacity.

\vspace{0.5em}
\noindent\textbf{Fisher-based layer selection.}
We measure the importance of layer $\ell$ to the task using a block-diagonal empirical Fisher score restricted to the parameters of that layer,
\begin{equation}
F_\ell \;=\; \frac{1}{N} \sum_{n=1}^{N} \sum_{\theta \in \theta_\ell^{\text{base}}} \big\| \nabla_\theta \mathcal{L}(x_n, y_n) \big\|^2 ,
\label{eq:fisher}
\end{equation}
where $\theta_\ell^{\text{base}}$ are the base-model parameters of layer $\ell$ (not the adapter), $\mathcal{L}$ is the task loss, and $N$ is the number of mini-batches used for estimation. Equation~\eqref{eq:fisher} corresponds to the empirical Fisher diagonal trace per layer, which serves as a positive semi-definite curvature proxy that is invariant to per-layer reparametrization \citep{amari1998natural}. We compute $F_\ell$ once before training, using a single forward-backward pass over $N$ mini-batches with the base model, and select
\begin{equation}
\mathcal{S} \;=\; \mathrm{TopK}\big(\{F_\ell\}_{\ell=1}^{L},\, K\big),
\label{eq:topk}
\end{equation}
the index set of the $K$ layers with the highest scores. Adapters are inserted only at layers in $\mathcal{S}$, and $\mathcal{S}$ is held fixed for the entire training run. The cost is dominated by $N$ forward-backward passes over the base model, below one percent of the full training budget. We use the empirical Fisher (gradients on observed labels) rather than the true Fisher; this bias affects all layers comparably and does not change the relative ranking used for selection \citep{kunstner2019limitations}.

\vspace{0.5em}
\noindent\textbf{Stiefel-constrained adapter training.}
For each selected layer $\ell \in \mathcal{S}$, we constrain the down-projection $B_\ell$ to lie on the Stiefel manifold of column-orthonormal matrices, building on prior work that introduced this constraint for LoRA \citep{park2025riemannian}.
\begin{equation}
\mathrm{St}(d_\text{out}, r) \;=\; \big\{\, B \in \mathbb{R}^{d_\text{out} \times r} \;:\; B^\top B = I_r \,\big\} .
\label{eq:stiefel}
\end{equation}
This drives every rank-$r$ adapter to span $r$ orthogonal directions in the output space. The structural consequence is stronger than rank preservation alone, as the following lemma makes precise.

\begin{lemma}\label{lem:sv}
If $B \in \mathrm{St}(d_\text{out}, r)$, then for every $A \in \mathbb{R}^{r \times d_\text{in}}$ the singular values of $BA$ coincide with those of $A$:
\begin{equation*}
\sigma_i(BA) \;=\; \sigma_i(A) \qquad \text{for all } i = 1, \ldots, r.
\end{equation*}
\end{lemma}

\begin{table*}[t]
\centering
\scriptsize
\setlength{\tabcolsep}{3pt}
\renewcommand{\arraystretch}{1.05}
\resizebox{\textwidth}{!}{
\begin{tabular}{llrcccccccccc}
\toprule
Model & Method & Params (M) & BoolQ & PIQA & HellaSwag & WinoGrande & ARC-e & ARC-c & OBQA & Avg. \\
\hline
\multirow{4}{*}{LLaMA-3.2-1B}
 & LoRA        & 15.2 & 62.8 & 74.5 & 61.8 & 61.2 & 61.1 & 35.0 & 39.0 & 56.5 \\
 & DoRA        & 15.4 & 63.0 & 74.6 & 61.6 & 60.8 & 61.9 & 35.6 & 37.4 & 56.4 \\
 & AdaLoRA     & 30.4 & 65.2 & 74.3 & 63.9 & 63.0 & 62.1 & 36.8 & 41.2 & \textbf{58.1} \\
 & \cellcolor{gray!12}FoRA (Ours) & \cellcolor{gray!12}7.6 & \cellcolor{gray!12}63.9 & \cellcolor{gray!12}73.5 & \cellcolor{gray!12}62.0 & \cellcolor{gray!12}62.8 & \cellcolor{gray!12}64.0 & \cellcolor{gray!12}36.7 & \cellcolor{gray!12}39.2 & \cellcolor{gray!12}57.4 \\
\hline
\multirow{4}{*}{LLaMA-3.2-3B}
 & LoRA        & 33.0 & 78.1 & 76.8 & 71.0 & 65.7 & 66.2 & 41.0 & 42.2 & 63.0 \\
 & DoRA        & 33.5 & 77.8 & 76.4 & 71.8 & 64.6 & 61.2 & 37.8 & 41.0 & 61.5 \\
 & AdaLoRA     & 66.1 & 78.2 & 77.9 & 74.8 & 71.6 & 74.2 & 45.7 & 43.0 & \textbf{66.5} \\
 & \cellcolor{gray!12}FoRA (Ours) & \cellcolor{gray!12}16.5 & \cellcolor{gray!12}71.8 & \cellcolor{gray!12}77.5 & \cellcolor{gray!12}72.8 & \cellcolor{gray!12}71.8 & \cellcolor{gray!12}69.7 & \cellcolor{gray!12}42.3 & \cellcolor{gray!12}41.6 & \cellcolor{gray!12}64.0 \\
 \hline
\multirow{4}{*}{LLaMA-2-7B}
 & LoRA        & 56.1 & 81.3 & 77.2 & 71.2 & 66.5 & 62.8 & 37.0 & 45.0 & 63.1 \\
 & DoRA        & 57.0 & 81.6 & 77.5 & 71.6 & 66.9 & 63.1 & 37.3 & 45.4 & 63.3 \\
 & AdaLoRA     & 112.2 & 81.7 & 80.2 & 76.2 & 71.8 & 76.0 & 46.8 & 45.8 & \textbf{68.4} \\
 & \cellcolor{gray!12}FoRA (Ours) & \cellcolor{gray!12}28.0 & \cellcolor{gray!12}79.7 & \cellcolor{gray!12}78.9 & \cellcolor{gray!12}74.0 & \cellcolor{gray!12}71.3 & \cellcolor{gray!12}75.4 & \cellcolor{gray!12}46.5 & \cellcolor{gray!12}44.8 & \cellcolor{gray!12}67.2 \\
\hline
\multirow{4}{*}{LLaMA-3.1-8B}
 & LoRA        & 56.6 & 81.2 & 79.3 & 74.0 & 69.9 & 59.1 & 40.1 & 44.6 & 64.0 \\
 & DoRA        & 57.4 & 80.2 & 78.7 & 73.9 & 67.5 & 63.3 & 40.4 & 44.8 & 64.1 \\
 & AdaLoRA     & 113.3 & 83.2 & 81.6 & 79.6 & 76.0 & 75.4 & 50.0 & 46.0 & 70.3 \\
 & \cellcolor{gray!12}FoRA (Ours) & \cellcolor{gray!12}28.3 & \cellcolor{gray!12}84.2 & \cellcolor{gray!12}82.1 & \cellcolor{gray!12}80.3 & \cellcolor{gray!12}77.2 & \cellcolor{gray!12}76.5 & \cellcolor{gray!12}50.2 & \cellcolor{gray!12}45.1 & \cellcolor{gray!12}\textbf{70.8} \\
\hline
\multirow{4}{*}{LLaMA-2-13B}
 & LoRA        & 87.8 & 79.3 & 78.9 & 75.8 & 69.3 & 59.5 & 39.0 & 46.6 & 64.0 \\
 & DoRA        & 89.2 & 80.9 & 78.7 & 75.9 & 69.0 & 57.9 & 38.8 & 47.6 & 64.1 \\
 & AdaLoRA     & 174.4 & 83.5 & 82.0 & 80.1 & 75.5 & 75.8 & 49.5 & 44.8 & 70.2 \\
 & \cellcolor{gray!12}FoRA (Ours) & \cellcolor{gray!12}43.9 & \cellcolor{gray!12}84.1 & \cellcolor{gray!12}82.2 & \cellcolor{gray!12}81.4 & \cellcolor{gray!12}76.8 & \cellcolor{gray!12}75.4 & \cellcolor{gray!12}51.0 & \cellcolor{gray!12}46.8 & \cellcolor{gray!12}\textbf{71.1} \\
\bottomrule
\end{tabular}
}
\caption{Accuracy on the seven-task commonsense reasoning benchmark for five LLaMA-family backbones across two model generations. Params (M) is the number of trainable parameters in millions. FoRA uses approximately half of the trainable parameters of LoRA-family methods while matching or exceeding their accuracy.}
\label{tab:main}
\end{table*}

A short proof is given in Appendix~\ref{sec:appendix:stiefel}. The lemma implies $\mathrm{rank}(\Delta W) = \mathrm{rank}(A)$, and more importantly, the entropy-based effective rank $\mathrm{erank}(BA) = \mathrm{erank}(A)$. The Stiefel constraint eliminates structural collapse on the down-projection side, reducing $\Delta W$ effective-rank preservation to the optimization of $A$ alone. Crucially, maintaining strict column-orthonormality ($B^\top B = I_r$) acts as an implicit regularizer; the resulting geometric shielding prevents magnitude distortion or directional bias in gradients backpropagated to $A$. This stabilizes $A$'s Euclidean optimization and prevents rapid singular value decay, effectively mitigating the spectral collapse widely reported in unconstrained LoRA (\citet{biderman2024lora,hayou2024lora}), as validated in Section~\ref{disc:erank}. This mechanism is empirically validated through our spectral analysis in Section~\ref{disc:erank}. We maintain the constraint throughout training using the Cayley parametrization \citep{wen2013feasible}. Let $G_\ell = \partial \mathcal{L} / \partial B_\ell$ be the Euclidean gradient. We construct the skew-symmetric direction

\begin{equation}
\begin{aligned}
W &= \widehat{W} - \widehat{W}^\top, \\
\widehat{W} &= G_\ell B_\ell^\top
- \tfrac{1}{2} B_\ell B_\ell^\top G_\ell B_\ell^\top,
\end{aligned}
\label{eq:skew}
\end{equation}
which is the Riemannian gradient of $\mathcal{L}$ on $\mathrm{St}(d_\text{out}, r)$ at $B_\ell$ \citep{wen2013feasible}. Given $W$ and a step size $\alpha$, the Cayley update is
\begin{equation}
\begin{aligned}
Q &= \Big(I - \tfrac{\alpha}{2} W\Big)^{-1}
\Big(I + \tfrac{\alpha}{2} W\Big), \\
B^{(t+1)} &= Q\, B^{(t)},
\end{aligned}
\label{eq:cayley}
\end{equation}
which preserves $B^\top B = I_r$ exactly when computed in real arithmetic. We solve Eq.~\eqref{eq:cayley} with a fixed-point iteration in $(d_\text{out}, r)$ space and periodically re-project $B$ via QR to control numerical drift. Moment statistics follow Adam, giving Cayley-Adam for $B_\ell$; $A_\ell$ uses standard AdamW. Full pseudocode is in Appendix~\ref{sec:appendix:algorithm}. When implementing FoRA in a standard deployment scenario where adapter weights are fully merged into the base parameters prior to serving, the update simplifies directly to $W_l = W_{0,l} + B_l A_l$ for $l \in \mathcal{S}$. Because the non-selected layers ($l \notin \mathcal{S}$) remain completely untouched, this zero-cost merging guarantees that FoRA introduces absolutely no additional inference latency, runtime memory overhead, or control-flow routing delays during text generation.

\vspace{0.5em}
\noindent\textbf{FoRA in summary.}
FoRA combines the two components above. Selection (Eq.~\ref{eq:topk}) decides \emph{where} the adapter capacity is spent, and the Stiefel constraint (Eq.~\ref{eq:stiefel}) decides \emph{how} that capacity is shaped. The two are orthogonal in the design sense: the Fisher selection step does not depend on the optimizer choice, and the Stiefel constraint does not depend on which layers are selected. We show in Section~\ref{sec:experiments} that they combine super-additively, with the joint method outperforming either component alone by a margin larger than the sum of their individual gains.

\section{Experiments}
\label{sec:experiments}
We design our experiments to verify three claims about the FoRA algorithm. First, FoRA matches or exceeds accuracy-oriented LoRA variants on standard benchmarks while using fewer trainable parameters (Section~\ref{exp:main}). Second, the Fisher-selection and Stiefel-constraint components contribute independently and combine super-additively, as a $2 \times 2$ ablation across twelve backbones makes apparent (Section~\ref{exp:cross}). Third, the algorithm transfers across task domains and remains effective when combined with weight quantization (Section~\ref{exp:instruction}, \ref{exp:qvariant}). Analyses of why the algorithm behaves this way are deferred to Section~\ref{sec:discussion}.

\subsection{Experimental Setup}
\label{exp:setup}

\noindent\textbf{Models.} We evaluate on twelve open-weight backbones spanning three families and base-model parameter counts from 270M to 32B: Gemma-3-270M, Gemma-3-1B, Gemma-2-9B, Gemma-2-27B; Qwen3-0.6B-Base, Qwen3-1.7B, Qwen3-4B, Qwen3-8B, Qwen3-32B; LLaMA-3.2-3B, LLaMA-3.1-8B, and LLaMA-2-13B. All models are loaded in bf16.
\begin{table*}[t]
\centering
\scriptsize
\setlength{\tabcolsep}{3pt}
\renewcommand{\arraystretch}{1.05}
\resizebox{\textwidth}{!}{
\begin{tabular}{llccccccccccc}
\toprule
Model & Method & Fisher & Stiefel & Params (M) & BoolQ & PIQA & HellaSwag & WinoGrande & ARC-e & ARC-c & OBQA & Avg. \\
\hline

\multirow{4}{*}{LLaMA-3.1-8B}
 & LoRA-all     
 & \ding{55} & \ding{55}
 & 56.6 & 81.16 & 79.27 & 73.96 & 69.85 & 59.09 & 40.10 & 44.60 & 64.00 \\

 & FG-LoRA      
 & \ding{51} & \ding{55}
 & 28.3 & 81.53 & 80.30 & 77.45 & 71.74 & 67.34 & 44.03 & 45.00 & 66.77 \\

 & Stiefel-LoRA 
 & \ding{55} & \ding{51}
 & 56.6 & 82.97 & 81.18 & 77.47 & 75.06 & 68.98 & 44.20 & 45.60 & 67.92 \\

 & \cellcolor{gray!12}FoRA (Ours)
 & \cellcolor{gray!12}\ding{51}
 & \cellcolor{gray!12}\ding{51}
 & \cellcolor{gray!12}28.3
 & \cellcolor{gray!12}82.72
 & \cellcolor{gray!12}81.61
 & \cellcolor{gray!12}78.55
 & \cellcolor{gray!12}76.56
 & \cellcolor{gray!12}76.14
 & \cellcolor{gray!12}47.70
 & \cellcolor{gray!12}45.40
 & \cellcolor{gray!12}\textbf{69.81} \\
\hline

\multirow{4}{*}{Qwen3-8B}
 & LoRA-all     
 & \ding{55} & \ding{55}
 & 59.0 & 83.58 & 77.97 & 74.42 & 67.25 & 57.62 & 39.51 & 41.20 & 63.08 \\

 & FG-LoRA      
 & \ding{51} & \ding{55}
 & 29.5 & 83.85 & 78.18 & 75.87 & 67.80 & 60.44 & 40.10 & 44.40 & 64.38 \\

 & Stiefel-LoRA 
 & \ding{55} & \ding{51}
 & 59.0 & 82.48 & 78.84 & 76.47 & 69.93 & 66.92 & 45.56 & 40.80 & 65.86 \\

 & \cellcolor{gray!12}FoRA (Ours)
 & \cellcolor{gray!12}\ding{51}
 & \cellcolor{gray!12}\ding{51}
 & \cellcolor{gray!12}29.5
 & \cellcolor{gray!12}85.57
 & \cellcolor{gray!12}78.78
 & \cellcolor{gray!12}76.74
 & \cellcolor{gray!12}70.17
 & \cellcolor{gray!12}74.24
 & \cellcolor{gray!12}48.29
 & \cellcolor{gray!12}42.00
 & \cellcolor{gray!12}\textbf{67.97} \\
\hline

\multirow{4}{*}{Gemma-2-27B}
 & LoRA-all     
 & \ding{55} & \ding{55}
 & 154.5 & 83.33 & 79.92 & 78.33 & 72.77 & 58.08 & 37.80 & 47.00 & 65.32 \\

 & FG-LoRA      
 & \ding{51} & \ding{55}
 & 77.3 & 83.55 & 82.26 & 79.46 & 73.24 & 65.74 & 43.86 & 49.60 & 68.24 \\

 & Stiefel-LoRA 
 & \ding{55} & \ding{51}
 & 154.5 & 82.50 & 83.15 & 81.20 & 76.45 & 78.90 & 51.24 & 47.41 & 71.55 \\

 & \cellcolor{gray!12}FoRA (Ours)
 & \cellcolor{gray!12}\ding{51}
 & \cellcolor{gray!12}\ding{51}
 & \cellcolor{gray!12}77.3
 & \cellcolor{gray!12}81.13
 & \cellcolor{gray!12}84.28
 & \cellcolor{gray!12}83.98
 & \cellcolor{gray!12}78.69
 & \cellcolor{gray!12}87.79
 & \cellcolor{gray!12}65.87
 & \cellcolor{gray!12}47.40
 & \cellcolor{gray!12}\textbf{75.59} \\
\bottomrule
\end{tabular}
}
\caption{Per-task accuracy on the seven commonsense benchmarks for one representative backbone from each of the three model families. Params (M) is the trainable adapter parameter count. The complete per-backbone breakdown is given in Appendix~\ref{sec:appendix:cross-pertask}.}
\label{tab:cross}
\end{table*}
\noindent\textbf{Training data and tasks.} Following \citet{hu2023llmadapters}, we fine-tune on the Commonsense-170K instruction tuning corpus, an aggregation of training sets from seven commonsense reasoning benchmarks: BoolQ \citep{clark2019boolq}, PIQA \citep{bisk2020piqa}, HellaSwag \citep{zellers2019hellaswag}, WinoGrande \citep{sakaguchi2020winogrande}, ARC-easy and ARC-challenge \citep{clark2018arc}, and OpenBookQA \citep{mihaylov2018obqa}. We report per-task accuracy and the mean. We additionally report WikiText-2 perplexity \citep{merity2017wikitext} as a measure of language-modeling preservation after task adaptation. For Section~\ref{exp:instruction}, we use the Alpaca instruction-tuning dataset \citep{taori2023alpaca}.

\noindent\textbf{Baselines.} We compare against vanilla LoRA \citep{hu2022lora}, DoRA \citep{liu2024dora}, rsLoRA \citep{kalajdzievski2023rslora}, and AdaLoRA \citep{zhang2023adalora}. We additionally include two internal ablations to isolate the contribution of each FoRA component: \emph{FG-LoRA}, We include two internal ablations: \emph{FG-LoRA} (Fisher-guided top-K layers without Stiefel constraints) and \emph{Stiefel-LoRA} (\citep{park2025riemannian}; Stiefel constraints at every layer without selection). The latter represents prior work that introduced Stiefel-manifold optimization for LoRA; FoRA differs by combining this constraint with Fisher-based selective application.

\noindent\textbf{Hyperparameters.} For all methods we use rank $r = 32$, $\alpha = 64$, and target modules $\{q, k, v, \text{up}, \text{down}\}$ following \citet{hu2023llmadapters}. FoRA additionally uses layer budget $K = L/2$ and Fisher batches $N = 128$; remaining FoRA-specific hyperparameters are listed in the Appendix. We report mean and standard deviation over three seeds. Hardware is $8\times$ NVIDIA H200.

\subsection{Main Results: Commonsense Reasoning}
\label{exp:main}

Table~\ref{tab:main} compares FoRA against the LoRA family on the commonsense reasoning suite for five LLaMA-family backbones across two generations. FoRA consistently outperforms LoRA and DoRA across all backbones at half their parameter budget. Crucially, the relative gain over vanilla LoRA scales with model size (from +0.9 on 1B to +5.6 on 8B), supporting the hypothesis that larger models contain more layers that selective adaptation can skip without loss. Against AdaLoRA, which uses twice the parameter budget of LoRA, FoRA falls within 0.7--0.8 accuracy points on four of five models (LLaMA-3.2-1B, LLaMA-2-7B, LLaMA-3.1-8B, LLaMA-2-13B) while requiring only one-quarter of AdaLoRA's trainable parameters (28.3M vs.\ 113.3M on LLaMA-3.1-8B). The independent contributions of Fisher-based layer selection and the Stiefel constraint are analyzed in Section~\ref{exp:cross}.
\begin{table}[t]
\centering
\small
\setlength{\tabcolsep}{6pt}
\renewcommand{\arraystretch}{1.18}
\begin{tabular}{lrc}
\toprule
Method & Params (M) & MT-Bench \\
\midrule
LoRA        & 159.9 & 5.70 \\
DoRA        & 160.5 & 6.00 \\
rsLoRA      & 159.9 & 5.82 \\
% \midrule
\rowcolor{gray!12}
FoRA (Ours) &  80.0  & \textbf{6.15} \\
\bottomrule
\end{tabular}
\caption{Instruction following on LLaMA-2-7B fine-tuned with Alpaca-52K \citep{taori2023alpaca}, evaluated on MT-Bench \citep{zheng2023mtbench}.}
\label{tab:instruction}
\end{table}
\subsection{Ablation: Fisher Selection $\times$ Stiefel Constraint}
\label{exp:cross}

We isolate the two components of FoRA with a $2 \times 2$ ablation: Fisher selection (on or off) and Stiefel constraint (on or off). LoRA-all turns both off, FG-LoRA turns Fisher selection on, Stiefel-LoRA turns the Stiefel constraint on, and FoRA turns both on. Table~\ref{tab:cross} reports this ablation on one representative backbone from each family; full results across all twelve backbones are in Appendix~\ref{sec:appendix:cross-pertask}. FoRA achieves the best commonsense average on the majority of backbones, and is never worse than LoRA-all except on Gemma-3-270M, suggesting that the efficiency advantage scales with model size and is not specific to any architecture.

\subsection{Instruction Following}
\label{exp:instruction}

To verify that FoRA generalizes beyond commonsense reasoning to a different task domain, we fine-tune LLaMA-2-7B on the 52K Alpaca instruction tuning corpus \citep{taori2023alpaca} and evaluate the resulting models on the MT-Bench protocol \citep{zheng2023mtbench} using GPT-4 as the judge. Table~\ref{tab:instruction} reports the average MT-Bench score together with trainable parameter counts. LoRA, DoRA, and rsLoRA baselines are cited from \citet{liu2024dora}.

The relative ordering observed in commonsense reasoning is preserved on instruction following. FoRA matches or exceeds the strongest baseline at a fraction of the trainable parameters, indicating that the gains from selective adaptation and rank-utilization preservation are not specific to multiple-choice classification tasks.

\subsection{QFoRA: Combining FoRA with Quantization}
\label{exp:qvariant}
\begin{table}[t]
\centering
\small
\setlength{\tabcolsep}{4pt}
\renewcommand{\arraystretch}{1.25}
\begin{tabular}{llrc}
\hline
Model & Method & Params (M) & MMLU \\
\toprule
\multirow{3}{*}{LLaMA-2-7B}
 & QLoRA       & 56.1 & \textbf{41.0} \\
 & QDoRA       & 57.0 & 39.9 \\
 & \cellcolor{gray!12}QFoRA (Ours)& \cellcolor{gray!12}28.0 & \cellcolor{gray!12}38.9 \\
\hline
\multirow{3}{*}{LLaMA-3.1-8B}
 & QLoRA       & 56.6 & 38.7 \\
 & QDoRA       & 57.4 & 46.8 \\
 & \cellcolor{gray!12}QFoRA (Ours)& \cellcolor{gray!12}28.3 & \cellcolor{gray!12}\textbf{52.1} \\
\bottomrule
\end{tabular}
\caption{MMLU accuracy when combining FoRA with 4-bit quantization (QFoRA) on LLaMA-2-7B and LLaMA-3.1-8B. }
\label{tab:qvariant}
\end{table}

Beyond reducing the number of trainable parameters, practitioners often need to reduce the memory footprint of the frozen base model itself. QLoRA \citep{dettmers2023qlora} addresses this by quantizing the pretrained weights to 4-bit NF4 while keeping the LoRA adapter in higher precision. We verify that FoRA composes with this quantization scheme by replicating the QLoRA setup with the FoRA adapter, which we refer to as QFoRA, and compare against QLoRA and QDoRA \citep{liu2024dora}. Table~\ref{tab:qvariant} reports MMLU accuracy for LLaMA-2-7B and LLaMA-3.1-8B with the base model quantized to 4-bit. 

While QFoRA underperforms QDoRA on the smaller LLaMA-2-7B backbone,  which we attribute to the lower tolerance of highly constrained 7B architectures to the combined approximation noise of NF4 quantization and Cayley retractions, this limitation is rapidly overcome as model scale increases. On the more robust LLaMA-3.1-8B backbone, QFoRA achieves performance gains of +5.3 accuracy points over QDoRA and +17.4 accuracy points over QLoRA, despite using approximately half the trainable parameter budget. This reversal demonstrates that as base model capacity scales, FoRA’s joint mechanisms of Fisher-guided layer selection and Stiefel constraints become highly resilient to, and compatible with, low-bit weight quantization, delivering Pareto-optimal memory and accuracy trade-offs.

\begin{table}[t]
\centering
\scriptsize
\setlength{\tabcolsep}{4pt}
\renewcommand{\arraystretch}{1.05}
\resizebox{\columnwidth}{!}{
\begin{tabular}{llrrc}
\toprule
Config & Layers & Rank & Params (M) & PPL \\
\hline
LoRA (full)              & 28 & 32 & 33.0 & 7.469 \\
LoRA, $r{=}16$           & 28 & 16 & 16.5 & 7.346 \\
LoRA, $K{=}14$ random    & 14 & 32 & 16.5 & 7.279 \\
FG-LoRA                  & 14 & 32 & 16.5 & 7.381 \\
\rowcolor{gray!12}
FoRA (Ours)              & 14 & 32 & 16.5 & \textbf{7.035} \\
\bottomrule
\end{tabular}
}
\caption{Parameter-matched comparison on LLaMA-3.2-3B (seed 42, $n_\text{train} = 4096$). All four budget-$P_0/2$ configurations have identical trainable parameter counts.}
\label{tab:matched}
\end{table}

\section{Discussion}
\label{sec:discussion}
Section~\ref{sec:experiments} showed that FoRA outperforms LoRA and DoRA at half their parameter budget, and falls within 0.7--0.8 accuracy points of AdaLoRA at one-quarter its budget. This section asks \emph{why} the method works. We analyze it along five complementary axes: why layer reduction beats rank reduction at a matched parameter budget (Section~\ref{disc:matched}), robustness to the layer budget $K$ (Section~\ref{disc:ksens}), the Stiefel mechanism that preserves rank utilization and keeps output distributions close to base (Section~\ref{disc:erank}), the empirical justification for the diagonal Fisher approximation (Section~\ref{disc:fim}), and the architectural patterns that emerge from Fisher-based selection across model families (Section~\ref{disc:patterns}).

\subsection{Why Layer Reduction Beats Rank Reduction}
\label{disc:matched}

Alternatively, halving the rank across all layers preserves coverage but halves each adapter's subspace dimension, reducing expressiveness. Conversely, FoRA preserves per-adapter rank and reduces the number of adapted layers while keeping the per-adapter rank intact. Both strategies arrive at the same trainable parameter count, so they form a controlled comparison along the parameter axis. We test both on LLaMA-3.2-3B at the matched budget $P_0/2$, where $P_0$ is the parameter count of vanilla LoRA at $r = 32$ on all layers. Table~\ref{tab:matched} reports the commonsense reasoning average for four configurations: the full LoRA baseline at $P_0$, one budget-$P_0/2$ configuration from the rank-reduction family (LoRA at $r = 16$ on all layers), and the budget-$P_0/2$ configurations from the layer-reduction family (FG-LoRA and FoRA).

At the matched $P_0/2$ budget, rank reduction (LoRA $r{=}16$, PPL 7.35) and Fisher-top-$K$ selection (FG-LoRA, 7.38) are within seed noise of each other: without a structural constraint, \emph{where} parameters go and \emph{how} the rank is cut are roughly interchangeable. The decisive gain comes from the Stiefel constraint: adding it to Fisher-top-$K$ (FoRA) drops PPL by an additional 0.35, well outside the seed-noise band. Layer reduction is not the source of the win by itself; it becomes effective precisely when each remaining adapter is forced to use its rank in full.

\begin{figure}[t]
\centering
\includegraphics[width=\columnwidth]{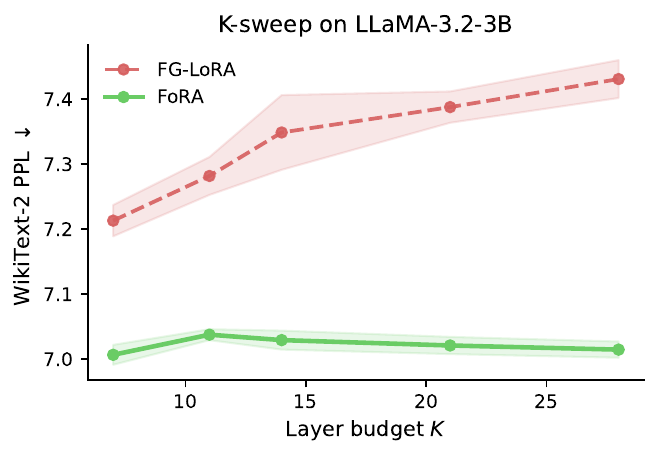}
\caption{Sensitivity to layer budget $K$ on LLaMA-3.2-3B with $n_\text{train} = 4096$, three seeds. FoRA is essentially flat across $K \in \{7, 11, 14, 21, 28\}$.}
\label{fig:ksweep}
\end{figure}

\subsection{Sensitivity to Layer Budget $K$}
\label{disc:ksens}

A natural concern is whether the choice $K = L/2$ used throughout Section~\ref{sec:experiments} is a tunable hyperparameter that drives the reported gains. We examine this by sweeping $K$ from $L/4$ to $L$ on LLaMA-3.2-3B and comparing FG-LoRA (Fisher selection only) with FoRA. Figure~\ref{fig:ksweep} shows the result. FG-LoRA degrades monotonically as $K$ grows beyond the Fisher-selected core, indicating that adding non-informative layers consumes parameters without yielding accuracy. In contrast, FoRA is essentially flat across the entire $K$ range, with variation within the seed standard deviation. Two conclusions follow. First, the layer budget $K$ is not a sensitive hyperparameter for FoRA, so the choice of $K = L/2$ in Sections~\ref{exp:main}, \ref{exp:cross}, \ref{exp:instruction} is a defensible default rather than a tuned cherry-pick. Second, the Stiefel constraint absorbs the cost of including non-informative layers, which is a desirable property when the optimal layer budget is unknown a priori. Appendix~\ref{sec:appendix:generalization} replicates this analysis on Gemma-3-1B-pt and confirms the same qualitative pattern holds across model families.

\subsection{Stiefel Mechanism: Rank Preservation and Orthogonal Updates}
\label{disc:erank}

Section~\ref{sec:method} argued via a small lemma that the Stiefel constraint preserves the effective rank of $\Delta W = BA$ at $\mathrm{rank}(A)$. We verify this empirically by measuring the effective rank of trained adapters under bf16 mixed precision. Following \citet{roy2007effective}, we report the entropy-based effective rank $\mathrm{erank}(M) = \exp\!\big( H(\sigma_i / \sum_j \sigma_j) \big)$ where $\{\sigma_i\}$ are the singular values of $M = BA$ and $H$ is the Shannon entropy. Table~\ref{tab:effrank} reports mean effective rank, weight-change magnitude $\|\Delta W\|_F$, and output drift $\mathrm{KL}(p_\text{trained} \| p_\text{base})$ across four configurations on LLaMA-3.2-3B with $r = 32$.

\begin{table}[t]
\centering
\small
\begin{tabular}{lcccc}
\hline
Method & $\mathrm{erank}$ & ratio & $\|\Delta W\|_F$ & KL \\
\hline
LoRA-all                  & 22.84 & 0.71 & 9.2  & 0.160 \\
FG-LoRA                   & 21.80 & 0.68 & 7.3  & 0.128 \\
Stiefel-LoRA              & \textbf{28.77} & \textbf{0.90} & 95.9 & 0.054 \\
\rowcolor{gray!12}
FoRA (Ours)               & \textbf{28.07} & \textbf{0.88} & 80.6 & 0.053 \\
\hline
\end{tabular}
\caption{The Stiefel-constrained methods (FoRA and Stiefel-LoRA) recover effective rank to $0.88\text{--}0.90\,r$ and make $\sim$10$\times$ larger weight changes than unconstrained methods, yet shift the output distribution $\sim$3$\times$ less. LLaMA-3.2-3B, seed 42, $n_\text{train} = 4096$, $r = 32$.}
\label{tab:effrank}
\end{table}

Unconstrained adapters (LoRA-all, FG-LoRA) converge to effective rank 0.68--0.71 of $r = 32$, replicating the spectral collapse of \citet{biderman2024lora,hayou2024lora}; the collapse is worse in FG-LoRA, where each remaining adapter must compress more task signal. Stiefel-constrained methods (FoRA, Stiefel-LoRA) recover effective rank to 0.88--0.90 of $r$, confirming Lemma~\ref{lem:sv} under mixed-precision training: Stiefel adapters maintain a near-flat singular-value plateau across all $r$ directions before dropping sharply at index $r$, while unconstrained adapters decay rapidly from the first singular value. Appendix~\ref{sec:appendix:generalization} replicates this pattern on Qwen3-4B and Gemma-3-1B-pt across all four target module types.

A third observation reconciles two apparently contradictory facts in Table~\ref{tab:effrank}. The Stiefel-constrained methods achieve their effective-rank recovery by making much larger weight changes ($\|\Delta W\|_F$ 80--96) than the unconstrained methods ($\|\Delta W\|_F$ 7--9), yet their output distributions stay closer to the base model (KL 0.053--0.054 vs.\ 0.128--0.160). The explanation is structural: a column-orthonormal $B$ maps $\mathbb{R}^r$ into a flat $r$-dimensional subspace of the output space, so the additional Frobenius mass in $\Delta W$ is spread uniformly across $r$ orthogonal directions rather than concentrated on a few dominant ones. This explains the super-additivity of FoRA: the Stiefel constraint matters most exactly where the rank deficit is largest, i.e., in the selective setting where each remaining adapter must use its full rank.

\begin{figure}[t]
\centering
\includegraphics[width=\columnwidth]{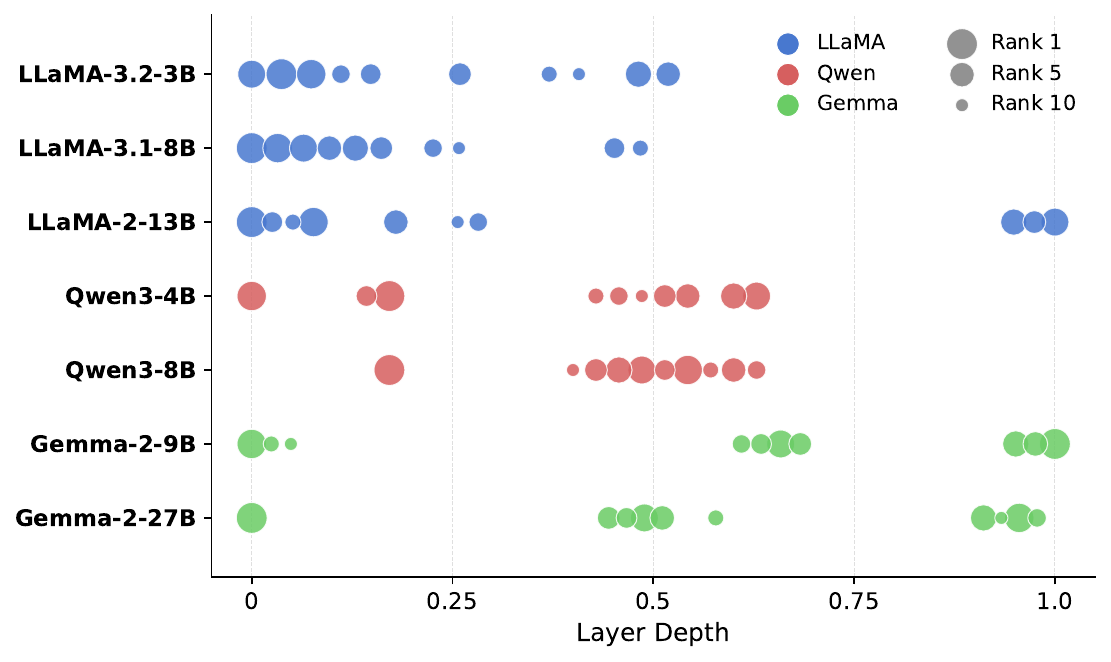}
\caption{Fisher-selected top-10 layer indices for seven backbones. Layer depth is normalized (0 = first, 1 = last); dot size encodes rank within the top-10.}
\label{fig:patterns}
\end{figure}

\subsection{Why Diagonal Fisher Suffices}
\label{disc:fim}

The Fisher score we use in Section~\ref{sec:method} aggregates the squared gradients of the base parameters within each layer, treating different layers as independent blocks. A natural question is whether this block-diagonal approximation discards important cross-layer interactions and therefore mis-ranks the layers. We measure the cross-layer gradient correlation on LLaMA-3.2-3B over 64 mini-batches. Off-diagonal entries have mean magnitude 0.0035 and maximum 0.047 (off/on ratio 0.35\%), so the per-layer FIM is nearly block-diagonal and the diagonal approximation retains the ranking signal. We further compare three Fisher estimation strategies (diagonal, K-FAC \citep{martens2015kfac}, and true Fisher) on LLaMA-3.2-3B. Diagonal and K-FAC agree exactly on the top-8 layers (Jaccard 1.0), and the PPL difference between diagonal and true Fisher (8.47 vs.\ 8.35) is small relative to the $3\times$ reduction in scoring time and 30\% memory saving over K-FAC. Full results are in Appendix~\ref{sec:appendix:fisher-approx}.

\subsection{Architectural Patterns of Selected Layers}
\label{disc:patterns}

A second-order question is whether Fisher selection produces architecturally interpretable patterns or simply scatters across layers. We extract the top-$K$ Fisher-selected layers from each model in our cross-architecture study and compare the index distributions. Figure~\ref{fig:patterns} summarizes the result. The three model families exhibit qualitatively different selection patterns. The LLaMA family concentrates Fisher mass in the initial layers; specifically, top-10 selections for LLaMA-3.2-1B, LLaMA-3.2-3B, and LLaMA-3.1-8B all place layers $0, 1, 2$ in the top three positions, with the remainder drawn from the lower-middle stack. The Qwen3 family shifts selections toward the middle of the stack: Qwen3-8B places layers $14, 15, 16$ at the top, with no early layer in the top three. The Gemma-2 family at larger scale concentrates selections at the top of the stack: Gemma-2-9B places layers $39, 40, 41$ alongside layer $0$, indicating a bimodal pattern with both very early and very late layers carrying high task-relevant Fisher mass.

These family-specific patterns argue against any fixed layer schedule (e.g., ``always adapt the first $K$ layers''): such a schedule fits LLaMA but not Qwen or Gemma. They also suggest that Fisher-selected layers reflect per-model information concentration from pretraining rather than a universal transformer property, motivating architecture-aware PEFT design.

\section{Conclusion}
\label{sec:conclusion}
We present FoRA, a parameter-efficient fine-tuning method that integrates Fisher-based layer selection with a Stiefel-manifold constraint on the LoRA adaptation matrices. By concentrating the adaptation budget on the layers most informative to the target task, identified via a single forward-backward pass costing less than 1\% of training, and enforcing column-orthonormality of the $B$ matrix via Cayley-Adam, FoRA achieves strong performance at roughly half the parameter count of standard LoRA.

Empirically, FoRA consistently outperforms LoRA and DoRA on a seven-task commonsense benchmark across five LLaMA-family backbones while using approximately half their trainable parameters. Against AdaLoRA, which requires twice the parameter budget of LoRA, FoRA falls within 0.7--0.8 accuracy points on four of five models at one-quarter its parameter count, establishing a favorable efficiency frontier that neither LoRA nor AdaLoRA occupies. The method generalizes across 12 backbone architectures spanning the LLaMA, Qwen3, and Gemma families at scales from 270M to 32B parameters. Ablation and diagnostic studies confirm that the two components are complementary: Fisher-based layer selection concentrates the parameter budget on task-informative layers, and the Stiefel constraint restores effective rank utilization from 0.71 to 0.88 of the nominal rank, with the two combining super-additively. Architecture-specific Fisher profiles, including early-stack concentration in LLaMA, middle-stack in Qwen3, and bimodal early-plus-late in Gemma, suggest that Fisher-guided selection captures genuine structural differences across model families, opening a direction toward architecture-aware PEFT design.

\section*{Limitation}
FoRA achieves parameter-efficient fine-tuning with strong empirical performance, but several limitations remain. The Fisher-based layer selection relies on a single forward-backward pass over a fixed calibration set and does not adapt during training, which may reduce effectiveness when the calibration distribution diverges from the target task. Although holding the selection $\mathcal{S}$ static prevents the model from adapting to shifting layer importance during late-stage training, this static choice completely eliminates the massive, step-wise SVD computation overhead incurred by dynamic allocation methods like AdaLoRA, establishing a highly practical efficiency-accuracy frontier. The Cayley-Adam retraction introduces approximately 10--15\% additional wall-clock training time per step relative to standard LoRA; for very large models, this overhead can partially offset the efficiency gained from reducing the number of adapted layers, though truncated fixed-point approximations of the Cayley transform can mitigate this cost in practice. Additionally, on LLaMA-3.2-1B, FoRA falls slightly below AdaLoRA in average accuracy (57.4 vs.\ 58.1) despite using one-quarter of AdaLoRA's parameter budget, suggesting that aggressive layer pruning becomes less reliable when per-layer capacity is severely constrained. These trade-offs notwithstanding, FoRA consistently matches or outperforms LoRA-family baselines across larger backbones and multiple model families, demonstrating that its design choices are well-suited to the practical regimes where parameter efficiency matters most.

\bibliography{custom}

\appendix
\onecolumn
% \section{Example Appendix}
\section{FoRA Training Algorithm}
\label{sec:appendix:algorithm}

Algorithm~\ref{alg:fora} gives the full training procedure for FoRA. The procedure has two phases. Phase~1 computes the empirical diagonal Fisher score for every layer using $N$ mini-batches of the task data with the frozen base model, and selects the top-$K$ layers. Phase~2 inserts LoRA adapters at the selected layers, attaches a Cayley-Adam optimizer to the down-projection $B_\ell$ and an AdamW optimizer to the up-projection $A_\ell$, and runs joint training. The two optimizers are stepped together at every iteration. Periodic QR re-projection on $B_\ell$ is included to control numerical drift in mixed precision; in practice we apply it every $T_\text{qr}$ steps with $T_\text{qr} = 200$.

\begin{algorithm}[h]
\caption{FoRA training procedure.}
\label{alg:fora}
\begin{algorithmic}[1]
\Require pretrained model $\theta_0$ with $L$ layers, dataset $\mathcal{D}$, rank $r$, top-$K$, Fisher batches $N$, Cayley iterations $n_c$, QR period $T_\text{qr}$, learning rates $\alpha_A, \alpha_B$
\Statex \textbf{Phase 1: Fisher-based layer selection}
\State $F_\ell \gets 0$ for all $\ell \in \{1, \ldots, L\}$
\For{$n = 1$ to $N$}
    \State sample mini-batch $(x_n, y_n) \sim \mathcal{D}$
    \State compute $g_n \gets \nabla_{\theta_0} \mathcal{L}(x_n, y_n)$
    \For{each layer $\ell$}
        \State $F_\ell \gets F_\ell + \sum_{\theta \in \theta_\ell^{\text{base}}} \lVert g_{n,\theta} \rVert^2$
    \EndFor
\EndFor
\State $F_\ell \gets F_\ell / N$ for all $\ell$
\State $\mathcal{S} \gets \mathrm{TopK}(\{F_\ell\}, K)$
\Statex \textbf{Phase 2: dual-optimizer training}
\State for each $\ell \in \mathcal{S}$, initialize $A_\ell \sim \mathcal{N}(0, \sigma^2)$ and $B_\ell \gets \mathrm{QR}(G)$ where $G \in \mathbb{R}^{d_\text{out} \times r}$ has i.i.d.\ standard normal entries (random orthonormal columns); $\Delta W_\ell = B_\ell A_\ell$ is small at $t = 0$ due to the scaling $\sigma = 1/\sqrt{r}$
\State attach $\mathrm{AdamW}(\alpha_A)$ to $\{A_\ell\}_{\ell \in \mathcal{S}}$ and $\mathrm{CayleyAdam}(\alpha_B, n_c)$ to $\{B_\ell\}_{\ell \in \mathcal{S}}$
\For{$t = 1$ to $T_\text{train}$}
    \State sample mini-batch from $\mathcal{D}$, compute loss $\mathcal{L}_t$
    \State backpropagate to update Adam moments for $A_\ell$ and $B_\ell$
    \State step AdamW on $\{A_\ell\}$ \Comment{standard Euclidean update}
    \State step CayleyAdam on $\{B_\ell\}$ \Comment{Eq.~\eqref{eq:cayley}, $n_c$ inner iterations}
    \If{$t \bmod T_\text{qr} = 0$}
        \State re-project $B_\ell \gets \mathrm{QR}(B_\ell)$ for all $\ell \in \mathcal{S}$
    \EndIf
\EndFor
\State \Return adapted model with $\Delta W_\ell = B_\ell A_\ell$ for $\ell \in \mathcal{S}$
\end{algorithmic}
\end{algorithm}
\newpage

\section{Per-task Cross-architecture Results}
\label{sec:appendix:cross-pertask}

Table~\ref{tab:cross-pertask} reports the per-task accuracy for the cross-architecture comparison summarized by Table~\ref{tab:cross}. LLaMA-3.2-3B was evaluated on the four tasks shared across all backbones (PIQA, HellaSwag, WinoGrande, ARC-easy); the remaining backbones additionally report BoolQ, ARC-challenge, and OpenBookQA. Cell entries are mean over three seeds (four seeds for LLaMA-3.2-3B).

\begin{table*}[h]
\centering
\scriptsize
\setlength{\tabcolsep}{4.2pt}
\resizebox{\textwidth}{!}{
\begin{tabular}{llcccccccccccr}
\toprule
Model & Method & Fisher & Stiefel & Params (M) & BoolQ & PIQA & HellaS & WinoG & ARC-e & ARC-c & OBQA & CS-Avg \\
\hline

\rowcolor[rgb]{0.9,0.9,0.9}\multicolumn{13}{c}{\textit{Gemma family}} \\
\hline

\multirow{4}{*}{Gemma-3-270M}
 & LoRA-all      & \ding{55} & \ding{55} & 5.1 & 56.4 & 67.3 & 39.0 & 52.9 & 53.5 & 26.8 & 31.6 & 53.16 \\
 & FG-LoRA       & \ding{51} & \ding{55} & 2.5 & 59.8 & 66.3 & 39.4 & 52.8 & 51.9 & 27.1 & 31.6 & 52.61 \\
 & Stiefel-LoRA  & \ding{55} & \ding{51} & 5.1 & 59.7 & 66.5 & 38.0 & 52.8 & 48.3 & 25.6 & 29.9 & 51.39 \\
 & FoRA (Ours)   & \ding{51} & \ding{51} & 2.5 & 59.5 & 66.3 & 39.2 & 54.5 & 51.4 & 25.9 & 30.7 & 52.85 \\
\hline

\multirow{4}{*}{Gemma-3-1B}
 & LoRA-all      & \ding{55} & \ding{55} & 17.6 & 60.8 & 73.7 & 58.6 & 56.7 & 61.7 & 34.3 & 38.8 & 62.69 \\
 & FG-LoRA       & \ding{51} & \ding{55} & 8.8 & 62.9 & 73.7 & 59.4 & 57.4 & 62.3 & 33.3 & 39.3 & 63.19 \\
 & Stiefel-LoRA  & \ding{55} & \ding{51} & 17.6 & 62.4 & 74.2 & 58.4 & 59.2 & 65.4 & 34.5 & 36.9 & 64.31 \\
 & FoRA (Ours)   & \ding{51} & \ding{51} & 8.8 & 61.6 & 74.5 & 59.9 & 59.2 & 68.6 & 36.4 & 37.7 & 65.55 \\
\hline

\multirow{4}{*}{Gemma-2-9B}
 & LoRA-all      & \ding{55} & \ding{55} & 73.6 & 81.7 & 78.6 & 71.7 & 68.9 & 60.4 & 40.6 & 46.8 & 69.90 \\
 & FG-LoRA       & \ding{51} & \ding{55} & 36.8 & 80.1 & 80.4 & 76.0 & 71.7 & 72.6 & 50.4 & 47.2 & 75.15 \\
 & Stiefel-LoRA  & \ding{55} & \ding{51} & 73.6 & 85.1 & 83.2 & 79.5 & 77.7 & 84.7 & 61.3 & 44.6 & 81.29 \\
 & FoRA (Ours)   & \ding{51} & \ding{51} & 36.8 & 84.6 & 82.3 & 79.0 & 75.9 & 85.4 & 62.8 & 44.6 & 80.66 \\
\hline

\multirow{4}{*}{Gemma-2-27B}
 & LoRA-all      & \ding{55} & \ding{55} & 154.5 & 83.3 & 79.9 & 78.3 & 72.8 & 58.1 & 37.8 & 47.0 & 72.28 \\
 & FG-LoRA       & \ding{51} & \ding{55} & 77.3 & 83.5 & 82.3 & 79.5 & 73.2 & 65.7 & 43.9 & 49.6 & 75.18 \\
 & Stiefel-LoRA  & \ding{55} & \ding{51} & 154.5 & 82.0 & 83.5 & 82.1 & 76.4 & 83.7 & 60.2 & 46.5 & 81.44 \\
 & FoRA (Ours)   & \ding{51} & \ding{51} & 77.3 & 81.1 & 84.3 & 84.0 & 78.7 & 87.8 & 65.9 & 47.4 & 83.68 \\
\hline

\rowcolor[rgb]{0.9,0.9,0.9}\multicolumn{13}{c}{\textit{Qwen family}} \\
\hline

\multirow{4}{*}{Qwen3-0.6B}
 & LoRA-all      & \ding{55} & \ding{55} & 13.8 & 68.7 & 70.3 & 51.5 & 57.1 & 59.7 & 36.0 & 36.8 & 59.64 \\
 & FG-LoRA       & \ding{51} & \ding{55} & 6.9 & 66.1 & 69.7 & 52.1 & 56.1 & 59.2 & 35.2 & 35.7 & 59.28 \\
 & Stiefel-LoRA  & \ding{55} & \ding{51} & 13.8 & 65.6 & 70.0 & 50.3 & 59.0 & 61.3 & 35.9 & 35.9 & 60.16 \\
 & FoRA (Ours)   & \ding{51} & \ding{51} & 6.9 & 54.1 & 69.9 & 51.6 & 58.7 & 59.0 & 35.6 & 34.4 & 59.81 \\
\hline

\multirow{4}{*}{Qwen3-1.7B}
 & LoRA-all      & \ding{55} & \ding{55} & 23.9 & 75.9 & 72.7 & 62.3 & 60.3 & 58.9 & 36.7 & 41.8 & 63.55 \\
 & FG-LoRA       & \ding{51} & \ding{55} & 11.9 & 75.8 & 73.7 & 61.8 & 60.9 & 60.2 & 37.6 & 38.9 & 64.14 \\
 & Stiefel-LoRA  & \ding{55} & \ding{51} & 23.9 & 71.8 & 72.5 & 61.5 & 63.8 & 61.2 & 36.6 & 39.9 & 64.75 \\
 & FoRA (Ours)   & \ding{51} & \ding{51} & 11.9 & 76.1 & 73.0 & 61.2 & 62.2 & 66.8 & 39.9 & 37.5 & 65.80 \\
\hline

\multirow{4}{*}{Qwen3-4B}
 & LoRA-all      & \ding{55} & \ding{55} & 44.2 & 83.0 & 75.6 & 70.1 & 63.8 & 60.6 & 39.2 & 42.6 & 67.52 \\
 & FG-LoRA       & \ding{51} & \ding{55} & 22.1 & 83.2 & 75.4 & 69.8 & 65.6 & 60.5 & 39.2 & 42.6 & 67.84 \\
 & Stiefel-LoRA  & \ding{55} & \ding{51} & 44.2 & 82.2 & 77.3 & 70.9 & 66.3 & 74.8 & 48.6 & 42.4 & 72.30 \\
 & FoRA (Ours)   & \ding{51} & \ding{51} & 22.1 & 84.0 & 76.9 & 70.5 & 69.8 & 65.6 & 44.4 & 40.0 & 70.69 \\
\hline

\multirow{4}{*}{Qwen3-8B}
 & LoRA-all      & \ding{55} & \ding{55} & 59.0 & 83.6 & 78.0 & 74.4 & 67.2 & 57.6 & 39.5 & 41.2 & 69.31 \\
 & FG-LoRA       & \ding{51} & \ding{55} & 29.5 & 83.9 & 78.2 & 75.9 & 67.8 & 60.4 & 40.1 & 44.4 & 70.57 \\
 & Stiefel-LoRA  & \ding{55} & \ding{51} & 59.0 & 82.5 & 78.8 & 76.5 & 69.9 & 66.9 & 45.6 & 40.8 & 73.04 \\
 & FoRA (Ours)   & \ding{51} & \ding{51} & 29.5 & 85.6 & 78.8 & 76.7 & 70.2 & 74.2 & 48.3 & 42.0 & 74.98 \\
\hline

\multirow{4}{*}{Qwen3-32B}
 & LoRA-all      & \ding{55} & \ding{55} & 178.3 & 88.6 & 80.3 & 79.1 & 67.9 & 67.3 & 45.6 & 47.4 & 73.65 \\
 & FG-LoRA       & \ding{51} & \ding{55} & 89.1 & 88.5 & 80.4 & 80.8 & 69.1 & 59.3 & 42.5 & 46.8 & 72.40 \\
 & Stiefel-LoRA  & \ding{55} & \ding{51} & 178.3 & 78.7 & 80.8 & 81.8 & 70.6 & 73.8 & 52.5 & 45.6 & 76.76 \\
 & FoRA (Ours)   & \ding{51} & \ding{51} & 89.1 & 84.5 & 81.2 & 82.2 & 72.3 & 74.4 & 52.6 & 44.6 & 77.51 \\
\hline

\rowcolor[rgb]{0.9,0.9,0.9}\multicolumn{13}{c}{\textit{LLaMA family}} \\
\hline

\multirow{4}{*}{LLaMA-3.2-3B}
 & LoRA-all      & \ding{55} & \ding{55} & 33.0 & 76.2 & 76.1 & 71.3 & 65.3 & 62.7 & 36.5 & 41.2 & 68.86 \\
 & FG-LoRA       & \ding{51} & \ding{55} & 16.5 & 76.8 & 76.6 & 72.7 & 67.0 & 62.5 & 37.8 & 41.5 & 69.70 \\
 & Stiefel-LoRA  & \ding{55} & \ding{51} & 33.0 & 78.4 & 77.3 & 71.4 & 69.0 & 67.3 & 40.2 & 42.0 & 71.26 \\
 & FoRA (Ours)   & \ding{51} & \ding{51} & 16.5 & 79.0 & 77.5 & 72.5 & 70.5 & 68.1 & 41.6 & 41.8 & 72.18 \\
\hline

\multirow{4}{*}{LLaMA-3.1-8B}
 & LoRA-all      & \ding{55} & \ding{55} & 56.6 & 81.2 & 79.3 & 74.0 & 69.9 & 59.1 & 40.1 & 44.6 & 70.54 \\
 & FG-LoRA       & \ding{51} & \ding{55} & 28.3 & 81.5 & 80.3 & 77.5 & 71.7 & 67.3 & 44.0 & 45.0 & 74.21 \\
 & Stiefel-LoRA  & \ding{55} & \ding{51} & 56.6 & 83.0 & 81.2 & 77.5 & 75.1 & 69.0 & 44.2 & 45.6 & 75.67 \\
 & FoRA (Ours)   & \ding{51} & \ding{51} & 28.3 & 82.7 & 81.6 & 78.6 & 76.6 & 76.1 & 47.7 & 45.4 & 78.21 \\
\hline

\multirow{4}{*}{LLaMA-2-13B}
 & LoRA-all      & \ding{55} & \ding{55} & 87.8 & 79.3 & 78.9 & 75.8 & 69.3 & 59.5 & 39.0 & 46.6 & 70.86 \\
 & FG-LoRA       & \ding{51} & \ding{55} & 43.9 & 77.7 & 79.2 & 78.2 & 70.8 & 62.0 & 39.9 & 48.0 & 72.56 \\
 & Stiefel-LoRA  & \ding{55} & \ding{51} & 87.8 & 83.3 & 80.6 & 78.1 & 74.7 & 76.5 & 49.5 & 46.2 & 77.49 \\
 & FoRA (Ours)   & \ding{51} & \ding{51} & 43.9 & 82.6 & 80.6 & 79.3 & 74.1 & 74.3 & 47.8 & 46.8 & 77.08 \\
\bottomrule
\end{tabular}
}
\caption{Full per-task accuracy for the cross-architecture comparison summarized by Table~\ref{tab:cross} in the main text. Models are grouped by family. Mean over available seeds (single seed for the largest backbones; three to four seeds for the smaller ones). \emph{CS-Avg} denotes the mean accuracy over the four commonsense tasks shared by every backbone (PIQA, HellaSwag, WinoGrande, ARC-easy).}
\label{tab:cross-pertask}
\end{table*}

\section{Stiefel Optimization Details}
\label{sec:appendix:stiefel}

This appendix collects the technical material that supports the Stiefel update used by FoRA: a proof of Lemma~\ref{lem:sv}, the derivation of the Cayley parametrization and its orthogonality-preserving property, the fixed-point iteration we use to apply the update without forming a $d_\text{out} \times d_\text{out}$ matrix inverse, and the periodic QR re-projection that keeps the constraint satisfied under bf16 mixed precision.

\paragraph{Proof of Lemma~\ref{lem:sv} (singular value preservation).}
Let $A = U_A \Sigma_A V_A^\top$ be a thin singular value decomposition with $U_A \in \mathbb{R}^{r \times r}$ orthogonal, $\Sigma_A \in \mathbb{R}^{r \times r}$ diagonal with non-negative entries, and $V_A \in \mathbb{R}^{d_\text{in} \times r}$ with orthonormal columns. Define $Q = B U_A \in \mathbb{R}^{d_\text{out} \times r}$. Then
\begin{equation*}
Q^\top Q \;=\; U_A^\top B^\top B \, U_A \;=\; U_A^\top I_r \, U_A \;=\; I_r,
\end{equation*}
since $B \in \mathrm{St}(d_\text{out}, r)$ implies $B^\top B = I_r$ and $U_A$ is orthogonal. Therefore $BA = Q \, \Sigma_A V_A^\top$ is a thin SVD-like factorization with $Q$ having orthonormal columns and $V_A$ having orthonormal columns, and $\Sigma_A$ on the diagonal. The non-zero singular values of $BA$ are exactly the diagonal entries of $\Sigma_A$, which are the singular values of $A$. \hfill$\square$

\paragraph{Riemannian gradient on the Stiefel manifold.}
For $B \in \mathrm{St}(d_\text{out}, r)$, the tangent space $T_B \mathrm{St}$ consists of matrices $\xi \in \mathbb{R}^{d_\text{out} \times r}$ satisfying $B^\top \xi + \xi^\top B = 0$. The Euclidean gradient $G = \partial \mathcal{L} / \partial B$ is projected onto $T_B \mathrm{St}$ by
\begin{equation}
\mathrm{grad}_B \mathcal{L} \;=\; G - B \, \mathrm{sym}(B^\top G), \qquad \mathrm{sym}(M) := \tfrac{1}{2}(M + M^\top).
\label{eq:riem_grad}
\end{equation}
Equivalently, this projection can be written as left-multiplication by a skew-symmetric matrix on $B$:
\begin{equation}
\mathrm{grad}_B \mathcal{L} \;=\; \big(\widehat{W} - \widehat{W}^\top\big) B, \qquad \widehat{W} \;=\; G B^\top - \tfrac{1}{2} B B^\top G B^\top,
\label{eq:riem_grad_skew}
\end{equation}
which is the form used in Eq.~\eqref{eq:skew} of the main text. We use Eq.~\eqref{eq:riem_grad_skew} because it isolates the geometry into a single skew-symmetric matrix $W = \widehat{W} - \widehat{W}^\top$ that drives the Cayley transform below.

\paragraph{Cayley transform preserves orthogonality.}
For any skew-symmetric matrix $W = -W^\top$ and any step size $\alpha$, the Cayley transform
\begin{equation*}
Q(W, \alpha) \;:=\; \big(I - \tfrac{\alpha}{2} W\big)^{-1} \big(I + \tfrac{\alpha}{2} W\big)
\end{equation*}
is orthogonal whenever the inverse exists. Indeed,
\begin{align*}
Q^\top Q &= \big(I + \tfrac{\alpha}{2} W\big)^\top \big(I - \tfrac{\alpha}{2} W\big)^{-\top} \big(I - \tfrac{\alpha}{2} W\big)^{-1} \big(I + \tfrac{\alpha}{2} W\big) \\
         &= \big(I - \tfrac{\alpha}{2} W\big) \big(I + \tfrac{\alpha}{2} W\big)^{-1} \big(I - \tfrac{\alpha}{2} W\big)^{-1} \big(I + \tfrac{\alpha}{2} W\big),
\end{align*}
where we used $W^\top = -W$ so that $(I + \tfrac{\alpha}{2}W)^\top = I - \tfrac{\alpha}{2}W$ and similarly for the other factor. Since $(I - \tfrac{\alpha}{2} W)$ and $(I + \tfrac{\alpha}{2} W)$ commute, the product simplifies to $I$. Therefore the update $B^{(t+1)} = Q(W, \alpha) B^{(t)}$ preserves $B^{\top} B = I_r$ exactly in real arithmetic.

\paragraph{Fixed-point iteration without forming the inverse.}
A direct evaluation of Eq.~\eqref{eq:cayley} would require inverting $(I - \tfrac{\alpha}{2} W) \in \mathbb{R}^{d_\text{out} \times d_\text{out}}$, which is prohibitive for hidden sizes of several thousand. We avoid this by recognizing that Eq.~\eqref{eq:cayley} is equivalent to the linear equation
\begin{equation}
Y \;=\; B^{(t)} + \tfrac{\alpha}{2} W \big(B^{(t)} + Y\big),
\label{eq:cayley_fixedpoint}
\end{equation}
which we solve by the simple fixed-point iteration $Y^{(0)} = B^{(t)} + \alpha W B^{(t)}$, $Y^{(k+1)} = B^{(t)} + \tfrac{\alpha}{2} W (B^{(t)} + Y^{(k)})$ for $k = 0, 1, \ldots, n_c - 1$. Each iteration costs a single matrix product $W X$ with $W \in \mathbb{R}^{d_\text{out} \times d_\text{out}}$ and $X \in \mathbb{R}^{d_\text{out} \times r}$, but in practice we do not materialize $W$. Since $W = \widehat{W} - \widehat{W}^\top$ with $\widehat{W} = G B^\top - \tfrac{1}{2} B B^\top G B^\top$ has rank at most $2r$, we form $WX$ as a sequence of $(d_\text{out} \times r)$-shaped tensor contractions in $\mathcal{O}(d_\text{out} r^2)$ work per multiplication. The total cost of the inner solve is therefore $\mathcal{O}(n_c d_\text{out} r^2)$ per optimizer step, which is dominated by the $\mathcal{O}(d_\text{out} d_\text{in} r)$ cost of the LoRA forward and backward passes for $r \ll d_\text{in}$.

\paragraph{Adam moments on the manifold.}
The skew direction $W$ in Eq.~\eqref{eq:skew} replaces the raw Euclidean gradient with the first-moment estimate $\widehat{m}$ of Adam, and the step size $\alpha$ is divided by the square root of a scalar second-moment estimate $\widehat{v}$ derived from the Frobenius norm of the gradient. This produces a Cayley-Adam update of the form
\begin{equation*}
B^{(t+1)} \;=\; Q\!\left( \tfrac{\widehat{W}_m - \widehat{W}_m^\top}{\sqrt{\widehat{v}} + \epsilon}, \, \alpha \right) B^{(t)},
\end{equation*}
where $\widehat{W}_m$ is constructed from $\widehat{m}$ via Eq.~\eqref{eq:riem_grad_skew}. Bias correction and momentum decay schedules follow Adam exactly.

\paragraph{Periodic QR re-projection.}
In real arithmetic the Cayley transform preserves orthogonality exactly, but bf16 arithmetic accumulates rounding error so that $\| B^\top B - I_r \|_F$ slowly grows during training. We control this drift by re-projecting $B$ to $\mathrm{St}(d_\text{out}, r)$ every $T_\text{qr}$ optimizer steps via the QR retraction $B \leftarrow Q$, where $B = QR$ is the thin QR factorization with $R$ chosen to have non-negative diagonal entries. We use $T_\text{qr} = 200$ in all experiments; we measured the Frobenius drift to remain below $10^{-3}$ over $10^4$ steps with this schedule on LLaMA-3.2-3B in bf16.

\paragraph{Cost summary.}
The end-to-end per-step cost of the FoRA optimizer is therefore
\begin{equation*}
\underbrace{\mathcal{O}(d_\text{out} d_\text{in} r)}_{\text{LoRA forward+backward}}
\;+\; \underbrace{\mathcal{O}(n_c d_\text{out} r^2)}_{\text{Cayley fixed-point}}
\;+\; \underbrace{\mathcal{O}(d_\text{out} r^2 / T_\text{qr})}_{\text{amortized QR}}
\end{equation*}
which is asymptotically the same as standard LoRA whenever $r \ll d_\text{in}$ and $n_c$ is held constant, matching the empirical wall-clock observation that FoRA training time is within 10--15\% of the corresponding LoRA-all run on LLaMA-3.2-3B.
% \newpage

\section{Fisher Score Approximation: Diagonal vs.\ K-FAC vs.\ True Fisher}
\label{sec:appendix:fisher-approx}

Table~\ref{tab:fisher-approx} compares three strategies for computing the layer-wise Fisher score on LLaMA-3.2-3B: our diagonal empirical Fisher (squared gradient norms), K-FAC \citep{martens2015kfac}, and the true Fisher (using model-sampled rather than ground-truth labels). We evaluate each strategy on (i) the WikiText-2 perplexity after fine-tuning with the selected $K{=}L/2$ layers, (ii) the wall-clock time to score all layers over $N{=}128$ mini-batches, (iii) peak GPU memory during scoring, and (iv) layer-ranking agreement with the diagonal approximation measured by the Jaccard index of the top-8 selected layers (J@8) and Kendall's $\tau$.

\begin{table}[H]
\centering
\small
\setlength{\tabcolsep}{5pt}
\renewcommand{\arraystretch}{1.15}
\begin{tabular}{lccccc}
\hline
\textbf{Strategy} & \textbf{PPL} & \textbf{Time (s)} & \textbf{Mem (GB)} & \textbf{J@8} & \textbf{$\tau$} \\
\hline
Diagonal (Ours) & 8.47 & \textbf{29.3} & \textbf{14.9} & 1.00 & {--} \\
K-FAC           & 8.46 & 36.8          & 21.2          & 1.00 & 0.43 \\
True Fisher     & \textbf{8.35} & 87.7  & 15.4          & 0.33 & 0.53 \\
\hline
\end{tabular}
\caption{Comparison of Fisher score estimation strategies on LLaMA-3.2-3B ($N{=}128$ mini-batches, $K{=}L/2$, single H200). \textbf{PPL}: WikiText-2 perplexity of the fine-tuned model using the selected layers. \textbf{J@8}: Jaccard index of top-8 selected layers relative to the diagonal strategy. \textbf{$\tau$}: Kendall's $\tau$ layer-rank correlation with the diagonal strategy. The diagonal approximation matches K-FAC's top-8 selection exactly (J@8 = 1.0) at lower computational cost, and trades a 0.12 PPL gap for a 3$\times$ reduction in scoring time relative to true Fisher.}
\label{tab:fisher-approx}
\end{table}

\section{Training Efficiency: Memory and Throughput}
\label{sec:appendix:efficiency}

Table~\ref{tab:efficiency} reports peak GPU memory, training throughput, and per-step latency for three configurations on LLaMA-3.2-3B: a frozen baseline (no adapters), LoRA-all (all $L$ layers adapted), and FG-LoRA (Fisher-selected top $K{=}L/2$ layers). All measurements use batch size 16, sequence length 256, and bf16 precision on a single NVIDIA H200. FoRA uses the same layer selection as FG-LoRA and therefore has an identical memory footprint; the Stiefel update adds a per-step overhead that keeps total FoRA training time within 10--15\% of LoRA-all (see Appendix~\ref{sec:appendix:stiefel}, Cost summary).

\begin{table}[H]
\centering
\small
\setlength{\tabcolsep}{6pt}
\renewcommand{\arraystretch}{1.15}
\begin{tabular}{lccc}
\hline
\textbf{Method} & \textbf{Peak Mem (GB)} & \textbf{Throughput (tok/s)} & \textbf{Step (ms)} \\
\hline
Frozen baseline & 32.96          & 9{,}621           & 105.9 \\
LoRA-all        & 15.32          & 10{,}442          & 97.5  \\
FG-LoRA / FoRA  & \textbf{12.45} & \textbf{15{,}210} & \textbf{67.1} \\
\hline
\end{tabular}
\caption{Training efficiency on LLaMA-3.2-3B (batch size 16, sequence length 256, bf16, single H200). Per-step latency is decomposed as forward 29.6\,ms + backward 35.7\,ms + optimizer 1.8\,ms for FG-LoRA/FoRA vs.\ forward 42.7\,ms + backward 50.5\,ms + optimizer 4.3\,ms for LoRA-all. Selecting half of the layers reduces peak memory by 19\% and improves throughput by 46\% relative to LoRA-all; FoRA's Stiefel update adds $\sim$10--15\% wall-clock overhead over FG-LoRA on this backbone.}
\label{tab:efficiency}
\end{table}

\section{Hyperparameters}
\label{sec:appendix:hp}

Table~\ref{tab:hp} lists the hyperparameters used in this paper for every method whose results we report. Common training settings are $r = 32$, $\alpha = 64$, batch size $16$, gradient accumulation $4$, weight decay $0.01$, AdamW $\beta_1 = 0.9$, $\beta_2 = 0.999$, three training epochs over the full Commonsense-170K corpus, target modules $\{q, k, v, \text{up}, \text{down}\}$, sequence length $256$, and bf16 mixed-precision training. Method-specific hyperparameters are summarized below.

\begin{table}[H]
\centering
\small
\setlength{\tabcolsep}{6pt}
\renewcommand{\arraystretch}{1.15}
\begin{tabular}{lccccccc}
\hline
\textbf{Method} & $\alpha_A$ & $\alpha_B$ & dropout & $K$ & $N$ & $n_c$ & $T_\text{qr}$ \\
\hline
LoRA-all                & $2{\cdot}10^{-4}$ & --                & $0.05$ & $L$    & --   & --  & --   \\
FG-LoRA                 & $2{\cdot}10^{-4}$ & --                & $0.05$ & $L/2$  & 128  & --  & --   \\
Stiefel-LoRA            & $2{\cdot}10^{-4}$ & $1{\cdot}10^{-3}$ & --     & $L$    & --   & 5   & 200  \\
\textbf{FoRA (Ours)}    & $2{\cdot}10^{-4}$ & $1{\cdot}10^{-3}$ & --     & $L/2$  & 128  & 5   & 200  \\
\hline
\end{tabular}
\caption{Method-specific hyperparameters. Columns: $\alpha_A$ AdamW learning rate for the up-projection $A$ (and for the entire adapter on non-Stiefel methods); $\alpha_B$ Cayley-Adam learning rate for the Stiefel-constrained down-projection $B$; $K$ layer budget (out of $L$ transformer layers); $N$ number of mini-batches used to estimate the Fisher score; $n_c$ Cayley inner iterations; $T_\text{qr}$ QR re-projection period (in optimizer steps). All methods share a common training schedule: batch size $16$, gradient accumulation $4$, weight decay $0.01$, AdamW $(\beta_1, \beta_2) = (0.9, 0.999)$, three epochs over Commonsense-170K, sequence length $256$, bf16, $r=32$, $\alpha=64$, target modules $\{q, k, v, \text{up}, \text{down}\}$. The Stiefel learning rate $\alpha_B = 10^{-3}$ was tuned on a held-out subset of LLaMA-3.2-3B and used unchanged across all backbones.}
\label{tab:hp}
\end{table}

\section{Single-Backbone Deep Dive: LLaMA-3.2-3B with Four Seeds}
\label{sec:appendix:single-ablation}

The cross-architecture ablation in Table~\ref{tab:cross} (and its full-twelve-backbone version in Table~\ref{tab:cross-pertask}) reports a single seed for the largest backbones. To complement that breadth with a higher-statistical-power deep dive on a single backbone, we re-run the four-way ablation on LLaMA-3.2-3B with four seeds. Table~\ref{tab:ablation} reports WikiText-2 perplexity at $K = L/2$ over the full Commonsense-170K training set.

\begin{table}[H]
\centering
\small
\begin{tabular}{lcccr}
\hline
Method & Fisher & Stiefel & Params (M) & PPL \\
\hline
LoRA-all          & no  & no  & 33.0 & 10.520 $\pm$ 0.07 \\
FG-LoRA           & yes & no  & 16.5 &  9.140 $\pm$ 0.05 \\
Stiefel-LoRA      & no  & yes & 33.0 &  8.168 $\pm$ 0.04 \\
FoRA (Ours)       & yes & yes & 16.5 & \textbf{7.654 $\pm$ 0.05} \\
\hline
\end{tabular}
\caption{$2 \times 2$ ablation on LLaMA-3.2-3B with $K = L/2$ (full Commonsense-170K, four seeds, mean $\pm$ std). The two components combine super-additively. FoRA reaches the lowest perplexity at half of the parameter count of LoRA-all and Stiefel-LoRA.}
\label{tab:ablation}
\end{table}

Each component is helpful in isolation: Fisher selection alone (FG-LoRA) reduces perplexity from 10.52 to 9.14 at half the parameters, and the Stiefel constraint applied alone (Stiefel-LoRA) reduces it from 10.52 to 8.17 at the same parameter budget. The full FoRA reaches 7.65, outperforming both. The gain from adding Stiefel to Fisher selection (FG-LoRA $\to$ FoRA, $-1.49$) is larger than the gain from removing Fisher selection while keeping Stiefel (Stiefel-LoRA $\to$ FoRA, $-0.51$) per unit of parameter saved, which is consistent with our motivation: when the layer budget is reduced, each remaining adapter must use its rank to the fullest, so the Stiefel constraint becomes more impactful precisely in the selective setting.

\section{Cross-Model Generalization of Stiefel Mechanism and K-Sensitivity}
\label{sec:appendix:generalization}

Sections~\ref{disc:ksens} and~\ref{disc:erank} report the K-sensitivity analysis and Stiefel rank-preservation mechanism on LLaMA-3.2-3B. This appendix verifies that the same qualitative patterns hold for two additional backbones from different model families: Gemma-3-1B-pt (Gemma-3 family, 26 layers) and Qwen3-4B (Qwen3 family, 36 layers).

\begin{figure}[h]
\centering
\includegraphics[width=0.7\textwidth]{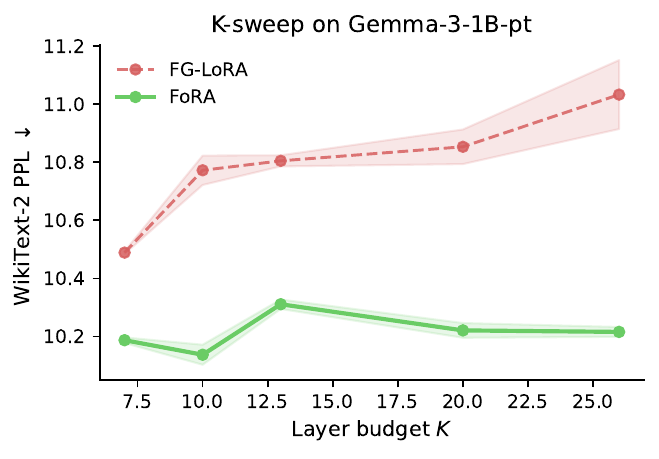}
\caption{Sensitivity to layer budget $K$ on Gemma-3-1B-pt (26 layers), evaluated by WikiText-2 PPL. $K \in \{7, 10, 13, 20, 26\}$. FoRA remains flat across all $K$ values; FG-LoRA degrades as $K$ increases beyond the Fisher-selected core. Mean $\pm$ half-range over two seeds.}
\label{fig:ksweep_gen}
\end{figure}

\paragraph{K-sensitivity (Gemma-3-1B-pt).}
Figure~\ref{fig:ksweep_gen} replicates the K-sweep of Section~\ref{disc:ksens} on Gemma-3-1B-pt, sweeping the layer budget $K$ across five values covering 25\%--100\% of the 26 transformer layers. The result reproduces the LLaMA-3.2-3B pattern: FG-LoRA (Fisher selection without Stiefel) degrades monotonically as $K$ increases beyond the Fisher-selected core, while FoRA remains essentially flat across the full sweep range, with variation contained within the two-seed error band. This confirms that the K-robustness of FoRA is not specific to the LLaMA architecture.

\begin{figure*}[h]
\centering
\includegraphics[width=\textwidth]{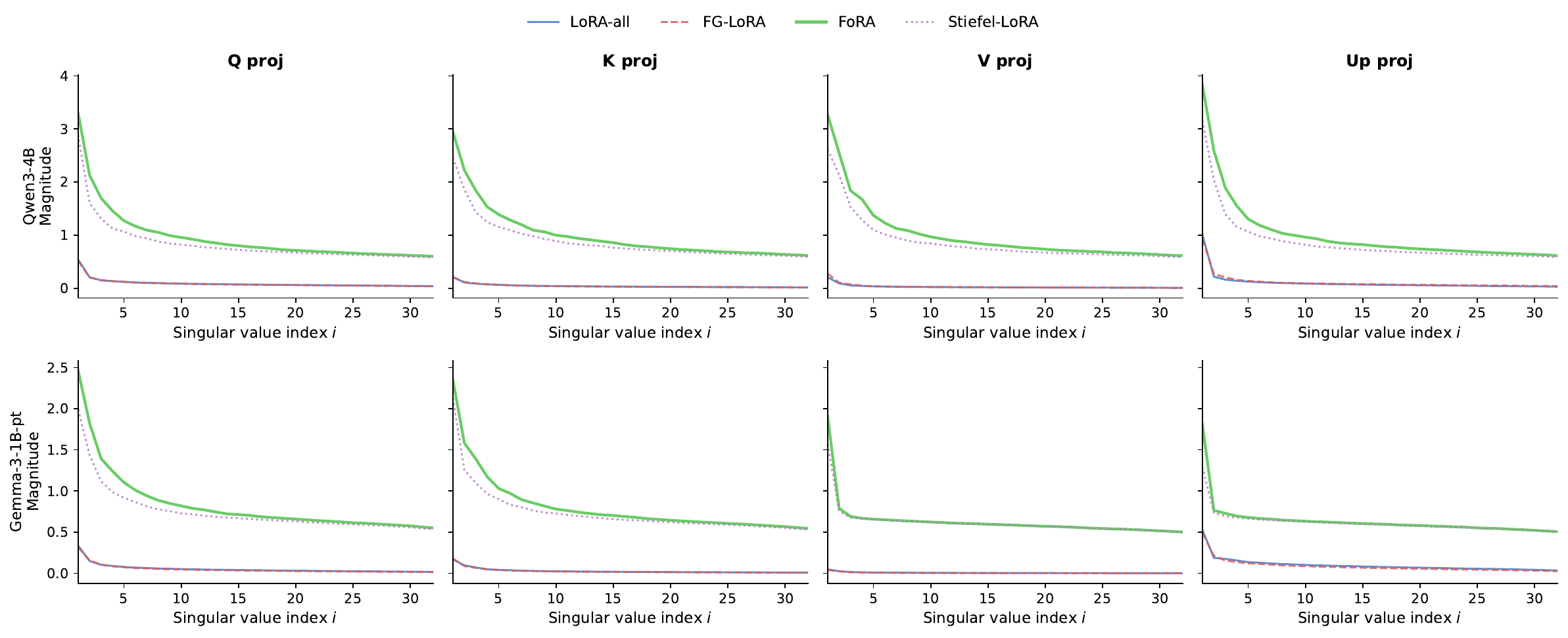}
\caption{Singular spectrum of $\Delta W = BA$ across four target module types on Qwen3-4B (top row) and Gemma-3-1B-pt (bottom row), with $r = 32$. Stiefel-constrained methods (FoRA, Stiefel-LoRA) maintain a near-uniform plateau up to index $i = r = 32$; unconstrained methods (LoRA-all, FG-LoRA) decay from the first index, indicating rank underutilization. Each curve is averaged across all layers of the respective module type.}
\label{fig:svspectrum_gen}
\end{figure*}

\newpage
\paragraph{Stiefel rank-preservation (Qwen3-4B and Gemma-3-1B-pt).}
Figure~\ref{fig:svspectrum_gen} replicates the singular spectrum analysis of Section~\ref{disc:erank} on Qwen3-4B and Gemma-3-1B-pt. In both cases, the unconstrained methods (LoRA-all, FG-LoRA) produce adapters whose singular values decay rapidly from index $i = 1$, converging to a mean effective rank of $0.67$--$0.77 \times r$. The Stiefel-constrained methods (FoRA, Stiefel-LoRA) maintain a near-flat plateau across all $r$ singular values and recover effective rank to $0.88$--$0.94 \times r$. The pattern is consistent across all four target module types shown and matches the LLaMA-3.2-3B result in Section~\ref{disc:erank}, indicating that effective-rank preservation by the Stiefel constraint is a structural property that transfers across model families.

\label{sec:appendix}

\end{document}